\documentclass[journal]{IEEEtran}
\IEEEoverridecommandlockouts

\usepackage{pgfplots}
\pgfplotsset{compat=newest}

\usepackage{enumitem}
\usepackage{array}
\usepackage{multirow}
\usepackage{rotating}
\usepackage{caption}
\usepackage{subcaption}
\usepackage{adjustbox}
\usepackage{cite}
\usepackage{hyperref}
\usepackage{amsmath,amssymb,amsfonts}
\usepackage{algorithmic}
\usepackage{graphicx}
\usepackage{textcomp}
\usepackage{xcolor}
\def\BibTeX{{\rm B\kern-.05em{\sc i\kern-.025em b}\kern-.08em
    T\kern-.1667em\lower.7ex\hbox{E}\kern-.125emX}}
    
\newcommand{\printfnsymbol}[1]{%
  \textsuperscript{*}%
}    

\begin{document}

\title{DeepSign: Deep On-Line Signature Verification
}

\author{Ruben Tolosana\printfnsymbol{1}\thanks{R. Tolosana and R. Vera-Rodriguez contributed equally to
this study.}, Ruben Vera-Rodriguez\printfnsymbol{1}\thanks{\copyright 2021 IEEE. Personal use of this material is permitted.  Permission from IEEE must be obtained for all other uses, in any current or future media, including reprinting/republishing this material for advertising or promotional purposes, creating new collective works, for resale or redistribution to servers or lists, or reuse of any copyrighted component of this work in other works.}, Julian Fierrez,~\IEEEmembership{Member,~IEEE}, Javier~Ortega-Garcia,~\IEEEmembership{Fellow,~IEEE} \\
Biometrics and Data Pattern Analytics - BiDA Lab, Universidad Autonoma de Madrid \\
{\tt\small \{ruben.tolosana, ruben.vera, julian.fierrez, javier.ortega\}@uam.es}
}


\maketitle

\begin{abstract}
Deep learning has become a breathtaking technology in the last years, overcoming traditional handcrafted approaches and even humans for many different tasks. However, in some tasks, such as the verification of handwritten signatures, the amount of publicly available data is scarce, what makes difficult to test the real limits of deep learning. In addition to the lack of public data, it is not easy to evaluate the improvements of novel proposed approaches as different databases and experimental protocols are usually considered. 

The main contributions of this study are: \textit{i)} we provide an in-depth analysis of state-of-the-art deep learning approaches for on-line signature verification, \textit{ii)} we present and describe the new DeepSignDB on-line handwritten signature biometric public database\footnote{\url{https://github.com/BiDAlab/DeepSignDB}}, \textit{iii)} we propose a standard experimental protocol and benchmark to be used for the research community in order to perform a fair comparison of novel approaches with the state of the art, and \textit{iv)} we adapt and evaluate our recent deep learning approach named Time-Aligned Recurrent Neural Networks (TA-RNNs)\footnote{Spanish Patent Application (P202030060)} for the task of on-line handwritten signature verification. This approach combines the potential of Dynamic Time Warping and Recurrent Neural Networks to train more robust systems against forgeries. Our proposed TA-RNN system outperforms the state of the art, achieving results even below 2.0\% EER when considering skilled forgery impostors and just one training signature per user.

\end{abstract}


\section{Introduction}
Handwritten signature verification is still an active research field nowadays~\cite{2020_Signature_CognitiveComputation}. Depending on the acquisition considered~\cite{moises_ACM}, it can be categorised as: \textit{i) off-line}, the signature is acquired in a traditional way by signing with an ink pen over paper and then digitizing the image; and \textit{ii) on-line}, the signature is acquired using electronic devices, having therefore not only the image of the signature, but also the signing information of the entire capturing process (time sequences).

On-line handwritten signature verification has widely evolved in the last 40 years~\cite{moises_ACM, eBioSign_journal}.  From the original Wacom devices specifically designed to acquire handwriting and signature in office-like scenarios to the current mobile acquisition scenarios in which signatures can be captured using our own personal smartphone anywhere. However, and despite the improvements achieved in the acquisition technology, the core of most of the state-of-the-art signature verification systems is still based on traditional approaches such as Dynamic Time Warping (DTW), Hidden Markov Models (HMM), and Support Vector Machines (SVM). This aspect seems to be a bit unusual if we compare with other biometric traits such as face and fingerprint in which deep learning has defeated by far traditional approaches~\cite{sundararajan2018deep,deepLearning_biometrics_vatsa,2019_TIFS_Fingerprint_PAs_Tolosana}, and even in tasks more related to signature verification such as handwriting recognition, writer verification, and handwritten passwords~\cite{online_offline_handwriting,end_to_end_writer_identification,2019_TMC_BioTouchPass_Tolosana}. So, why  deep learning approaches are not widely used in on-line signature verification yet? One major handicap could be probably the complex procedure of acquiring a large-scale database for training the models as signatures are not publicly available on internet as it happens with other biometric traits such as the face~\cite{kemelmacher2016megaface}.

\begin{figure*}[tb]
\centering
\centerline{\includegraphics[width=0.71\linewidth]{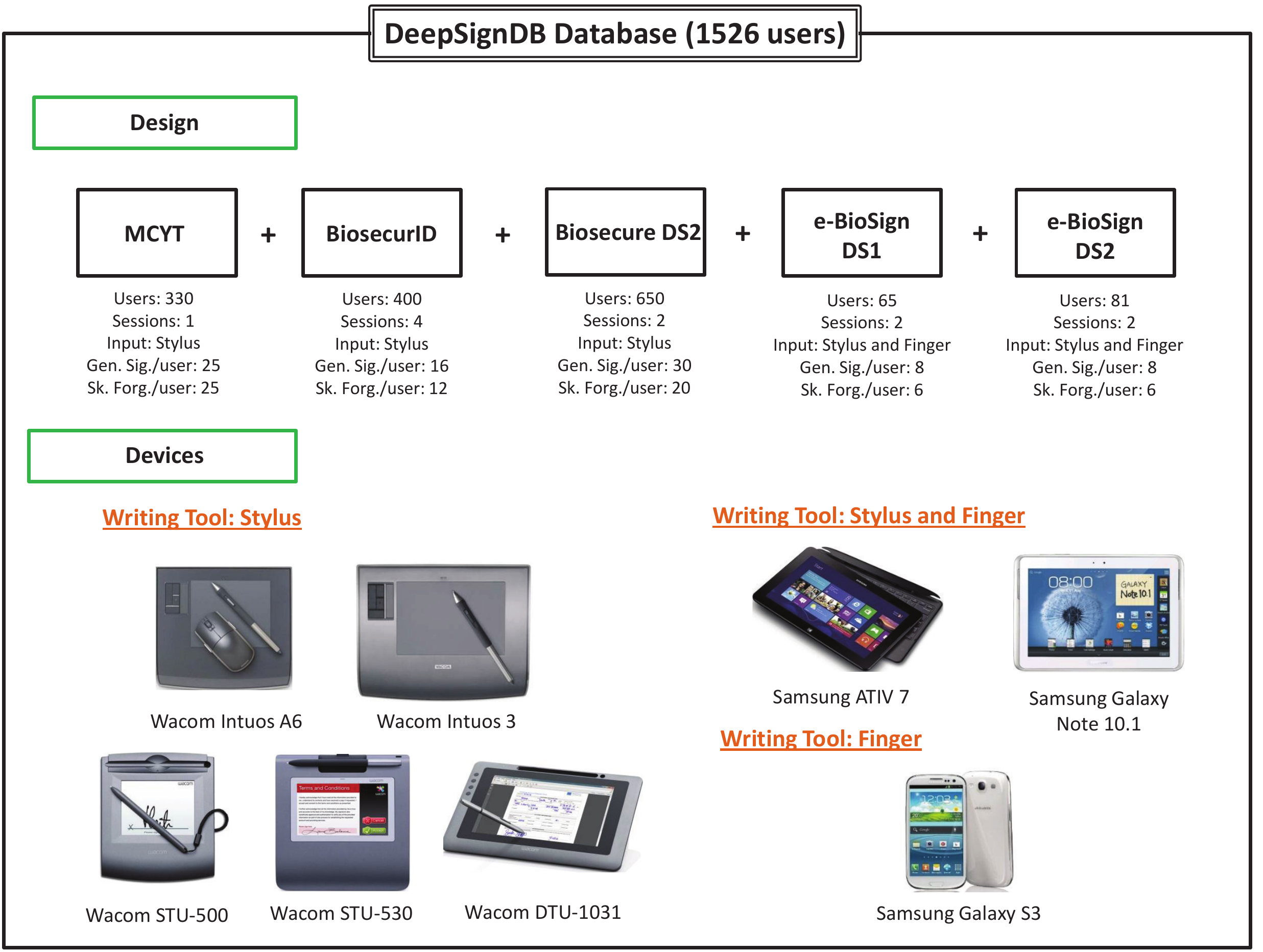}}
\caption{Description of the design, acquisition devices, and writing tools considered in the new DeepSignDB database. A total of 1526 users and 8 different captured devices are used (5 Wacom and 3 Samsung general-purpose devices). For the Samsung devices, signatures are also collected using the finger. Gen. Sig. = Genuine Signatures, and Sk. Forg. = Skilled Forgeries.} \label{fig:DeepSign_abstract}
\end{figure*}

\begin{figure*}[!]
\begin{center}
   \includegraphics[width=\linewidth]{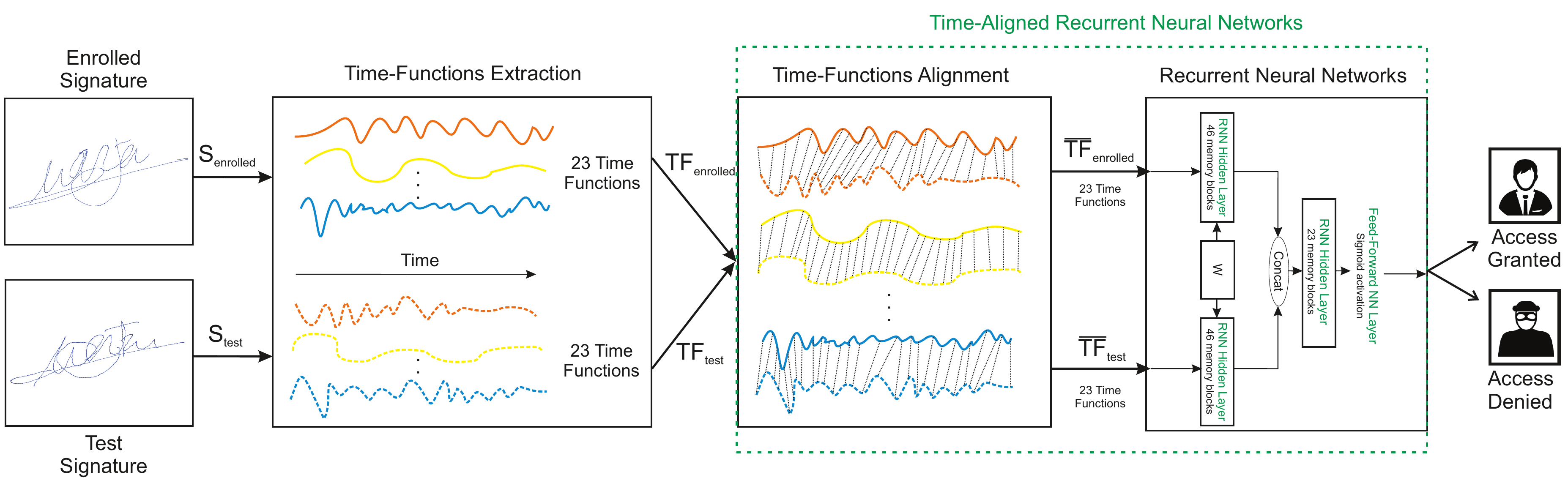}
\end{center}
   \caption{Architecture of our proposed on-line signature verification system based on Time-Aligned Recurrent Neural Networks. $S$ denotes one signature sample, and $TF$ and $\overline{TF}$ the original and pre-aligned 23 time functions, respectively. The Recurrent Neural Networks block is enlarged in Fig.~\ref{fig:LSTM_configuration_digits} for a better understanding. }
\label{fig:diagrama_TA-RNNs}
\end{figure*}

In addition to the scarcity of data for training deep learning approaches, another important observation motivates this work: the lack of a standard experimental protocol to be used for the research community in order to perform a fair comparison of novel approaches to the state of the art, as different experimental protocols and conditions are usually considered for different signature databases~\cite{diaz2018dynamic,liu2015online}. With all these concerns in mind, in this study we present the new DeepSignDB handwritten signature biometric database, the largest on-line signature database to date. Fig.~\ref{fig:DeepSign_abstract} graphically summarises the design, acquisition devices, and writing tools considered in the DeepSignDB database. Its application extends from the improvement of signature verification systems via deep learning to many other potential research lines, e.g., studying: \textit{i)} user-dependent effects, and development of user-dependent methods in signature biometrics, and handwriting recognition at large~\cite{yager2010biometric}, \textit{ii)} the neuromotor processes involved in signature biometrics~\cite{2019_BookLogNormal_ComplexSignTouch_Vera}, and handwriting in general~\cite{ferrer2018idelog}, \textit{iii)} sensing factors in obtaining representative and clean handwriting and touch interaction signals~\cite{2015_IEEEAccess_InterSign_Tolosana,Alonso-Fernandez2005_TabletPC}, \textit{iv)} human-device interaction factors involving handwriting and touchscreen signals~\cite{2019_TMC_BioTouchPass_Tolosana}, and development of improved interaction methods~\cite{harbach2016anatomy}, and \textit{v)} population statistics around handwriting and touch interaction signals, and development of new methods aimed at recognising or serving particular population groups~\cite{2018_IETB_DetectChildTouch_Acien,Vera_2019_Complexity}. 

The main contributions of this study are:
\begin{itemize}
\item An in-depth analysis of state-of-the-art deep learning approaches for on-line signature verification, remarking the different experimental protocol conditions considered among them.

\item The new DeepSignDB on-line handwritten signature database. This database is obtained through the combination of some of the most well-known databases, and a novel dataset not presented yet. It comprises more than 70K signatures acquired using both stylus and finger inputs from a total 1526 users. Two acquisition scenarios are considered, office and mobile, with a total of 8 different devices. Additionally, different types of impostors and number of acquisition sessions are considered.     

\item A standard experimental protocol publicly available to the research community in order to perform a fair comparison of novel approaches with the state of the art. Thus, we also release the files with all the signature comparisons carried out using the final evaluation dataset. This way we provide an easily reproducible framework.

\item An adaptation and evaluation of our recent deep learning approach named Time-Aligned Recurrent Neural Networks (TA-RNNs) for on-line handwritten signature verification. This approach was originally presented in~\cite{2020_TIFS_BioTouchPass2_Tolosana} for touchscreen password biometrics. Fig.~\ref{fig:diagrama_TA-RNNs} represents the general architecture of our proposed approach. It combines the potential of Dynamic Time Warping and Recurrent Neural Networks (RNNs) to train more robust systems against forgeries.

\item A benchmark evaluation of DeepSignDB considering well-known systems based on DTW, RNNs, and our newly proposed TA-RNNs. 
\end{itemize}

A preliminary version of this article was published in~\cite{Tolosana_2019_DeepSignDB_ICDAR}. This article significantly improves~\cite{Tolosana_2019_DeepSignDB_ICDAR} in the following aspects: \textit{i)} we provide an in-depth analysis of state-of-the-art deep learning approaches for on-line signature verification, \textit{ii)} we adapt and evaluate our recent TA-RNN deep learning approach, \textit{iii)} we provide a more extensive evaluation of DeepSignDB, analysing the system performance for each scenario and dataset of DeepSignDB and also for DTW, RNNs, and our proposed TA-RNNs, and \textit{iv)} our proposed TA-RNN approach further outperforms previous signature verification approaches, remarking the importance of time-functions alignment. 

The remainder of the paper is organised as follows. Sec.~\ref{DL_relatedWorks} summarises previous studies carried out in on-line signature verification via deep learning. Sec.~\ref{proposed_Approach} explains all details of our proposed TA-RNN approach. Sec.~\ref{sec_database_description} describes the details of the DeepSignDB signature database. Sec.~\ref{sec_deepSign_benchmark} describes the proposed experimental protocol, and the benchmark evaluation carried out. Finally, Sec.~\ref{conclusions} draws the final conclusions and points out some lines for future work.

\begin{table*}[t]
\centering
\caption{Comparison of different deep learning approaches for on-line signature verification.}
\label{table:data_relatedWorks}
\scalebox{0.85}{
\begin{tabular}{cccccccc}
\textbf{Study}                            & \textbf{Classifiers}              & \multicolumn{2}{c}{\textbf{Database}}                                 & \multicolumn{3}{c}{\textbf{Experimental Protocol}}                                                                           & \multicolumn{1}{l}{\textbf{Performance (EER)}}                        \\
\textbf{}                                 &                                   & \textbf{Name}                        & \textbf{\# Users}              & \textbf{\# Train Users}        & \textbf{Input}                   & \textbf{\# Train Sig.}                                   & \textbf{}                                                                          \\ \hline \hline
\begin{tabular}[c]{@{}c@{}}Otte \textit{et al.} (2014) \\ \cite{LSTM_signature_Liwicki}\end{tabular}                      & LSTM                              & SigComp2011                          & 20                             & 20                             & Stylus                           & 12                                                       & Skilled = 23.8\%                                                                   \\ \hline
\multirow{2}{*}{\begin{tabular}[c]{@{}c@{}}Tolosana \textit{et al.} (2018) \\ ~\cite{2018_IEEEAccess_RNN_Tolosana}\end{tabular}} & \multirow{2}{*}{BLSTM/BGRU}       & \multirow{2}{*}{BiosecurID}          & \multirow{2}{*}{400}           & \multirow{2}{*}{300}           & \multirow{2}{*}{Stylus}          & 1                                                        & \begin{tabular}[c]{@{}c@{}}Skilled = 6.8\%\\ Random = 5.4\%\end{tabular}           \\
                                          &                                   &                                      &                                &                                &                                  & 4                                                        & \begin{tabular}[c]{@{}c@{}}Skilled = 5.5\%\\ Random. = 2.9\%\end{tabular}          \\ \hline
\multirow{3}{*}{\begin{tabular}[c]{@{}c@{}}Lai and Jin (2018)\\ \cite{Lai_TIFS_2018}\end{tabular}}     & \multirow{3}{*}{GARU + DTW}             & MCYT                                 & 100                            & 80                             & Stylus                           & 5                                                        & \begin{tabular}[c]{@{}c@{}}Skilled = 1.8\%\\ Random = 0.2\%\end{tabular}           \\ \cline{3-8} 
                                          &                                   & Mobisig                              & 83                             & 70                             & Finger                           & 5                                                        & \begin{tabular}[c]{@{}c@{}}Skilled = 10.9\%\\ Random = 0.6\%\end{tabular}          \\ \cline{3-8} 
                                          &                                   & e-BioSign                            & 65                             & 30                             & Stylus                           & 4 & \begin{tabular}[c]{@{}c@{}}Skilled = 6.9\%\\ Random = 0.4\%\end{tabular}           \\ \hline

                                          \begin{tabular}[c]{@{}c@{}}Ahrabian and Babaali (2018) \\ \cite{ahrabian2018usage}\end{tabular}             & LSTM Autoencoder                  & SigWiComp2013                        & 31                             & 11                             & Stylus                           & 5                                                        & \begin{tabular}[c]{@{}c@{}}Skilled = 8.7\%\\ Random = Unknown\end{tabular} \\ \hline

                                          \begin{tabular}[c]{@{}c@{}}Hefny and Moustafa (2019) \\ \cite{hefny2019online}\end{tabular}             & MLP                  & SigComp2011                        & 64                             & -                             & Stylus                           & 5                                                        & \begin{tabular}[c]{@{}c@{}}Skilled = 0.5\%\\ Random = Unknown\end{tabular} \\ \hline

                                          \begin{tabular}[c]{@{}c@{}}Wu \textit{et al.} (2019) \\ \cite{Uchida2019_DDTW}\end{tabular}             & CNNs + DTW                  & MCYT                        & 100                             & 50                             & Stylus                           & 5                                                        & \begin{tabular}[c]{@{}c@{}}Skilled = 2.4\%\\ Random = Unknown\end{tabular} \\ \hline

\multirow{4}{*}{\begin{tabular}[c]{@{}c@{}}Li \textit{et al.} (2019)\\ \cite{Li2019_stroke_deepLearning}\end{tabular}}     & \multirow{4}{*}{LSTM}             & BiosecurID                                 & 132                            & 110                             & Stylus                           & 1                                                        & \begin{tabular}[c]{@{}c@{}}Skilled = 3.7\%\\ Random = 1.9\%\end{tabular}           \\ \cline{3-8} 
                                          &                                   & MCYT                              & 100                             & 85                             & Stylus                           & 1                                                        & \begin{tabular}[c]{@{}c@{}}Skilled = 10.5\%\\ Random = Unknown\end{tabular}          \\ \cline{3-8} 
                                          &                                   & SCUT-MMSIG                            & 50                             & 40                             & Stylus                           & 1 & \begin{tabular}[c]{@{}c@{}}Skilled = 13.9\%\\ Random = Unknown\end{tabular}  \\ \cline{3-8} 
                                          &                                   & Mobisig                            & 83                             & 70                             & Finger                           & 1 & \begin{tabular}[c]{@{}c@{}}Skilled = 16.1\%\\ Random = Unknown\end{tabular}         \\ \hline

                                            \multirow{3}{*}{\begin{tabular}[c]{@{}c@{}}Sekhar \textit{et al.} (2019)\\ \cite{chandra2019online}\end{tabular}}     & \multirow{3}{*}{CNNs}             & MCYT                                 & 100                            & 95                             & Stylus                           & 1                                                        & \begin{tabular}[c]{@{}c@{}}Skilled = 93.9\% Acc.\\ Random = Unknown\end{tabular}           \\ \cline{3-8} 
                                          &                                   & SVC-Task 2                              & 40                             & 35 & Stylus                           & 1                                                        & \begin{tabular}[c]{@{}c@{}}Skilled = 77.0\% Acc.\\ Random = Unknown\end{tabular}          \\ \hline

                                          \multirow{3}{*}{\begin{tabular}[c]{@{}c@{}}Lai \textit{et al.} (2020)\\ \cite{lai2020synsig2vec}\end{tabular}}     & \multirow{3}{*}{CNNs}             & MCYT                                 & 100                            & 90                             & Stylus                           & 5                                                        & \begin{tabular}[c]{@{}c@{}}Skilled = 1.7\%\\ Random = Unknown\end{tabular}           \\ \cline{3-8} 
                                          &                                   & SVC-Task 2                              & 40                             & 36                             & Stylus                           & 5                                                        & \begin{tabular}[c]{@{}c@{}}Skilled = 4.6\%\\ Random = Unknown\end{tabular}          \\ \hline

                                          \begin{tabular}[c]{@{}c@{}}Nathwani (2020) \\ \cite{nathwani2020online}\end{tabular}             & BLSTM/BGRU                  & SVC                        & -                             & -                             & Stylus                           & -                                                        & \begin{tabular}[c]{@{}c@{}}Skilled = 8.8\% AE\\ Random = Unknown\end{tabular} \\ \hline

\multirow{4}{*}{\textbf{Proposed}}        & \multirow{4}{*}{\textbf{TA-RNNs}} & \multirow{4}{*}{\textbf{DeepSignDB}} & \multirow{4}{*}{\textbf{1526}} & \multirow{4}{*}{\textbf{1084}} & \multirow{2}{*}{\textbf{Stylus}} & \textbf{1}                                               & \textbf{\begin{tabular}[c]{@{}c@{}}Skilled = 4.2\%\\ Random = 1.5\%\end{tabular}}  \\
                                          &                                   &                                      &                                &                                &                                  & \textbf{4}                                               & \textbf{\begin{tabular}[c]{@{}c@{}}Skilled = 3.3\%\\ Random = 0.6\%\end{tabular}}  \\ \cline{6-8} 
                                          &                                   &                                      &                                &                                & \multirow{2}{*}{\textbf{Finger}} & \textbf{1}                                               & \textbf{\begin{tabular}[c]{@{}c@{}}Skilled = 13.8\%\\ Random = 1.8\%\end{tabular}} \\
                                          &                                   &                                      &                                &                                &                                  & \textbf{4}                                               & \textbf{\begin{tabular}[c]{@{}c@{}}Skilled = 11.3\%\\ Random = 1.0\%\end{tabular}} \\ \hline
\end{tabular}
}
\end{table*}

\section{On-Line Signature Verification \\ Via Deep Learning }\label{DL_relatedWorks}

Despite the lack of publicly available data, some authors have preliminary evaluated the potential of different deep learning architectures for on-line signature verification. Table~\ref{table:data_relatedWorks} shows a comparison of different deep learning approaches with the corresponding database, experimental protocol, and performance results achieved. First, we would like to highlight the impossibility of performing a fair comparison among approaches as different databases and experimental protocol conditions have been considered in each study. Aspects such as the inter-session variability, the number of training signatures available per user or the complexity of the signatures have a very significant impact in the system performance~\cite{2019_IETB_Aging_Tolosana,tolosana2019exploiting}.  This problem is not only related to deep learning approaches, but to the whole handwritten signature verification field. 

One of the first studies that analysed the potential of current deep learning approaches for on-line signature verification was~\cite{LSTM_signature_Liwicki}. In that work, Otte \textit{et al.} performed an exhaustive analysis of Long Short-Term Memory (LSTM) RNNs using a total of 20 users and 12 genuine signatures per user for training. Three different scenarios were studied: \textit{i)} training a general network to distinguish forgeries from genuine signatures, \textit{ii)} training a different network for each writer, and \textit{iii)} training the network using only genuine signatures. However, all experiments failed obtaining a final 23.8\% EER for the best network configuration, far away from the state of the art, concluding that LSTM RNN systems trained with standard mechanisms were not appropriate for the task of signature verification as the amount of available data for this task is scarce compared with others, e.g., handwriting recognition. 

More recently, some researchers have preliminary shown the potential of deep learning for the task of on-line signature verification through the design of new architectures. In~\cite{2018_IEEEAccess_RNN_Tolosana}, the authors proposed an end-to-end writer-independent RNN signature verification system based on a Siamese architecture~\cite{chopra2005learning}. Both LSTM and Gated Recurrent Unit (GRU) schemes were studied, using both normal and bidirectional configurations (i.e., BLSTM and BGRU) in order to have access both to past and future context. The proposed system was able to outperform a state-of-the-art signature verification system based on DTW and feature selection techniques for the case of skilled forgeries. However, it was not able to outperform DTW for the case of random forgeries. 

Lai and Jin proposed in~\cite{Lai_TIFS_2018} the use of Gated Auto Regressive Units (GARU) in combination with a novel descriptor named Length-Normalized Path Signature (LNPS) in order to extract robust features. DTW was considered for the final classification. Experiments were carried out using different databases and experimental protocols, achieving good results especially against random forgeries. It is important to remark the results obtained using the Mobisig database with the finger as writing tool~\cite{antal2018online}. Their proposed approach achieved a final 10.9\% EER for skilled forgeries, much worse than the result achieved for MCYT database~\cite{Ortega_Garcia2003_MCYT}. This result highlights the challenging finger input scenario for signature verification~\cite{eBioSign_journal}. 

In this research line, in~\cite{ahrabian2018usage} the authors proposed a system based on an LSTM autoencoder for modelling each signature into a fixed-length feature latent space and a Siamese network for the final classification. The authors evaluated their approach over the SigWiComp2013 dataset~\cite{malik2013icdar} achieving around 8.7\% EER for skilled forgeries. 

Simpler approaches based on Multilayer Perceptron (MLP) were considered in~\cite{hefny2019online}. Hefny and Moustafa considered Legendre polynomials coefficients as features to model the signatures. Their proposed approach was tested using SigComp2011 (Dutch dataset)~\cite{liwicki2011signature}, achieving an EER of 0.5\%.

More recently, different authors have proposed novel approaches in ICDAR 2019\footnote{\url{https://icdar2019.org/}}. Approaches based on the combination of Convolutional Neural Networks (CNNs) and DTW were presented in~\cite{Uchida2019_DDTW}. Their proposed approach was tested only against skilled forgeries over the MCYT database~\cite{Ortega_Garcia2003_MCYT}, showing how the system performance is highly affected by the amount of training data. 

Also, Li \textit{et al.} proposed in~\cite{Li2019_stroke_deepLearning} a stroke-based LSTM system. Their proposed approach seemed to outperform the results achieved in~\cite{2018_IEEEAccess_RNN_Tolosana} for the BiosecurID database~\cite{Fierrez2009_PAA}. However, the results achieved in other databases were much worse, above 10\% EER, showing the poor generalisation capacity of the network.  

Similar to the approach presented in~\cite{2018_IEEEAccess_RNN_Tolosana}, Sekhar \textit{et al.} presented in ICDAR 2019 a Siamese CNN architecture. Their proposed approach was evaluated over the MCYT and SVC databases~\cite{Ortega_Garcia2003_MCYT,yeung2004svc2004}, achieving very different accuracies for each database. 

An interesting analysis using a lightweight one-dimensional CNN signature verification system was recently proposed in~\cite{lai2020synsig2vec}, using fixed-length representations from signatures of variable length. In addition, they studied the potential of synthesis techniques to eliminate the need of skilled forgeries during training. Their proposed approach was evaluated using MCYT and SVC databases~\cite{Ortega_Garcia2003_MCYT,yeung2004svc2004}, achieving good results against skilled forgeries.

Nathwani proposed in~\cite{nathwani2020online} an on-line signature verification based on BLSTM/BGRU. No much information regarding the system, architecture, and training procedure is provided in the paper. The best result achieved on SVC was an Average Error (AE) of 8.8\%.

Finally, we include in Table~\ref{table:data_relatedWorks} the results achieved using our proposed TA-RNN system over the new DeepSignDB database. Due to all the limitations highlighted, in this study we propose and release to the research community a standard experimental protocol for on-line signature verification with the aim to make possible future comparative analysis of new proposed architectures.

\section{TA-RNN Signature Verification System}\label{proposed_Approach}
This section describes our proposed Time-Aligned Recurrent Neural Networks for on-line signature verification. A graphical representation is included in Fig.~\ref{fig:diagrama_TA-RNNs}.

\subsection{Time-Functions Extraction}\label{subsec_FeatureExtractor}
Our proposed on-line signature verification system is based on time functions. For each signature acquired (i.e., $S_{enrolled}$ and $S_{test}$ in Fig.~\ref{fig:diagrama_TA-RNNs}), signals related to $X$ and $Y$ spatial coordinates and pressure are used to extract a set of 23 time functions (i.e., $TF_{enrolled}$ and $TF_{test}$ in Fig.~\ref{fig:diagrama_TA-RNNs}), following the same approach described in~\cite{Marcos08a}. Table~\ref{tabla:tablaLocalFeatures} provides a description of the 23 time functions considered in this study. Finally, time functions are normalised to keep them in the same range of values using the mean and standard deviation~\cite{2015_IEEEAccess_InterSign_Tolosana}.

\subsection{Time-Functions Alignment}\label{subsec_SWDTW}
One crucial point when comparing the similarity among time sequences is the proper alignment of them prior to calculating the similarity score through distance measurement functions (e.g., the Euclidean distance). DTW is one of the most popular algorithms in the literature, in particular for signature biometrics~\cite{fischer2017signature,malik2015icdar2015,diaz2018dynamic,iet:/content/journals/10.1049/iet-bmt.2013.0044}. The goal of DTW is to find the optimal warping path of a pair of time sequences $A$ and $B$ that minimises a given distance measure $d(A,B)$.

In our proposed approach, DTW is applied in a first stage in order to convert the 23 original time functions (i.e., $TF_{enrolled}$ and $TF_{test}$ in Fig.~\ref{fig:diagrama_TA-RNNs}) into 23 pre-aligned time functions (i.e., $\overline{TF}_{enrolled}$ and $\overline{TF}_{test}$ in Fig.~\ref{fig:diagrama_TA-RNNs}) before introducing them to the RNNs. This way our proposed RNN system is able to extract more meaningful features as all time sequences have been previously normalised jointly through the optimal warping path.

\begin{table}[tb]
\centering
\caption{Set of time functions considered in this study.}
\begin{adjustbox}{width=0.455\textwidth}
\begin{tabular}{p{0.7cm} p{6.8cm}}
\hline
\# & Feature \\
\hline \hline
1 & \textit{X}-coordinate: $x_n$  \\
\hline
2 & \textit{Y}-coordinate: $y_n$  \\
\hline
3 & Pen-pressure: $z_n$ \\
\hline
4 & Path-tangent angle: $\theta_n$  \\
\hline
5 & Path velocity magnitude: $v_n$ \\
\hline
6 & Log curvature radius: $\rho_n$ \\
\hline
7 & Total acceleration magnitude: $a_n$ \\
\hline
8-14 & First-order derivative of features 1-7: $\dot{x_n},\dot{y_n},\dot{z_n},\dot{\theta_n},\dot{v_n},\dot{\rho_n},\dot{a_n}$ \\
\hline
15-16 & Second-order derivative of features 1-2: $\ddot{x_n},\ddot{y_n}$ \\
\hline
17 & Ratio of the minimum over the maximum speed over a 5-samples window: $v^r_n$ \\
\hline
18-19 & Angle of consecutive samples and first order difference: $\alpha_n$, $\dot{\alpha_n}$ \\
\hline
20 & Sine: $s_n$ \\
\hline
21 & Cosine: $c_n$ \\
\hline
22 & Stroke length to width ratio over a 5-samples window: $r^5_n$ \\
\hline
23 & Stroke length to width ratio over a 7-samples window: $r^7_n$ \\
\hline
\end{tabular}
\end{adjustbox}
\label{tabla:tablaLocalFeatures}
\end{table}

\subsection{Recurrent Neural Networks}\label{subsec_SRNNs}
\begin{figure}[t]
\begin{center}
   \includegraphics[width=0.92\linewidth]{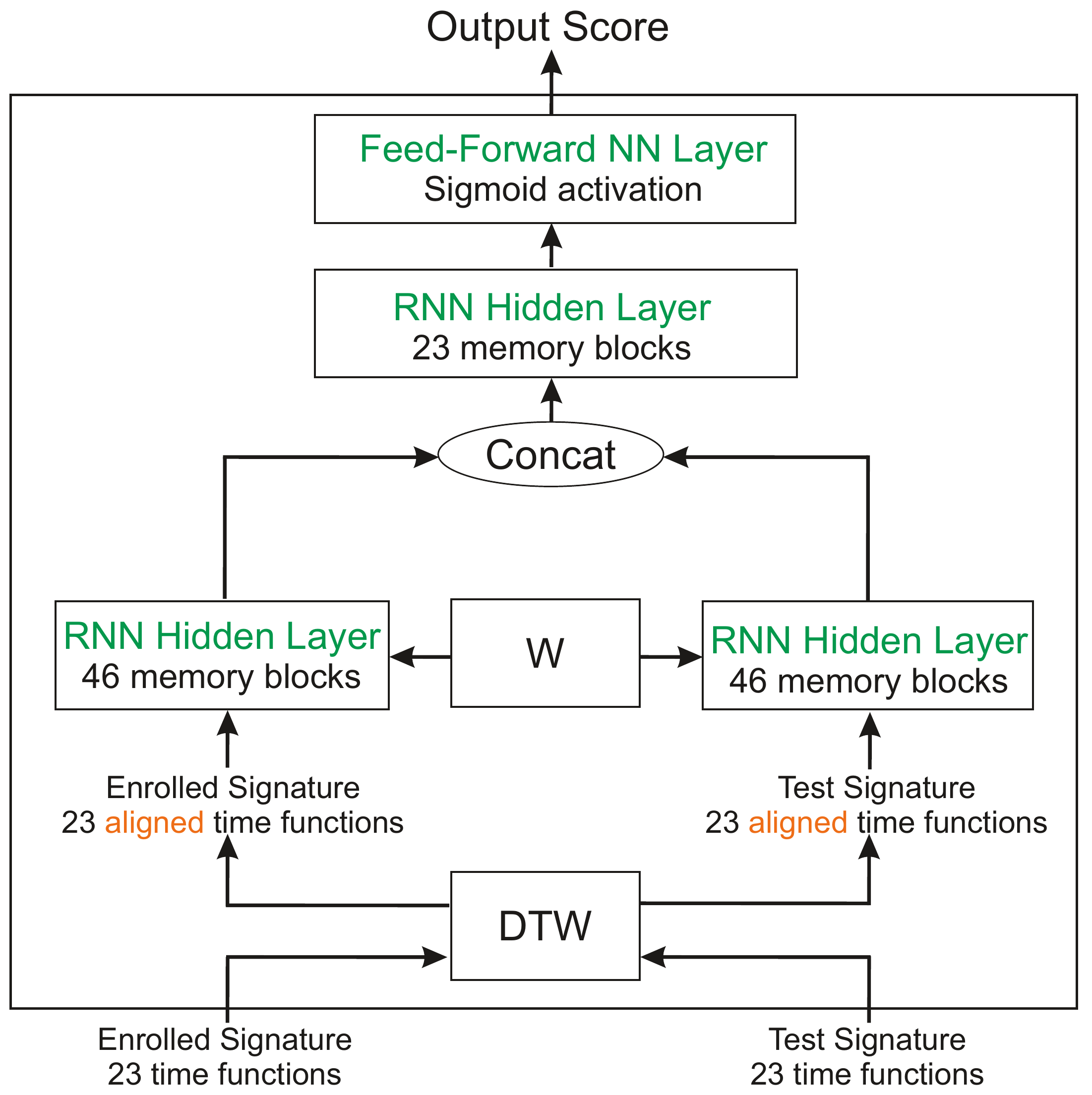}
\end{center}
   \caption{Proposed TA-RNN architecture.}
\label{fig:LSTM_configuration_digits}
\end{figure}

New trends based on the use of RNNs, which is a specific neural network architecture, are becoming more and more important nowadays for modelling sequential data with arbitrary length~\cite{yu2019review}. Fig.~\ref{fig:LSTM_configuration_digits} depicts our proposed TA-RNN system based on a Siamese architecture. The main goal is to learn a dissimilarity metric from data by minimising a discriminative cost function that drives the dissimilarity metric to be small for pairs of genuine signatures from the same subject (labelled as 0), and higher for pairs of genuine-forgery signatures (labelled as 1 for both random and skilled forgeries). This architecture is very similar compared with the initial one proposed in~\cite{2018_IEEEAccess_RNN_Tolosana} with the exception of the first stage based on time sequences alignment through DTW. Several configurations of the deep learning model were tested, changing the number of hidden layers and memory blocks as we did in~\cite{2018_IEEEAccess_RNN_Tolosana}, describing here the one that achieved the best results.

For the input of the network, we consider as much information as possible, i.e., all 23 time functions per signature previously aligned through DTW. Preliminary experiments suggested that it is better to feed the system with all time functions, letting the network to automatically select the more discriminative features on each epoch~\cite{2018_IEEEAccess_RNN_Tolosana}. The first layer is composed of two BGRU hidden layers with 46 memory blocks each, sharing the weights between them. The outputs of the first two parallel BGRU hidden layers are concatenated and serve as input to the second layer, which corresponds to a BGRU hidden layer with 23 memory blocks. The output of this second BGRU hidden layer is the vector resulting of the last timestep. Finally, a feed-forward neural network layer with a sigmoid activation is considered, providing an output score for each pair of signatures. It is important to highlight that our approach is trained to distinguish between genuine and impostor patterns from the signatures. Thus, we just train one writer-independent system for all databases through the development dataset.

\begin{table*}[t]
\centering
\caption{Specifications of the acquisition devices considered in each dataset of DeepSignDB.}
\label{table:devices_features}
\begin{tabular}{cccccc}
\textbf{Database}      & \textbf{Device}          & \textbf{Input} & \textbf{Screen (Diagonal)}              & \textbf{Sensor Resolution} & \textbf{Sampling Rate} \\ \hline
\textbf{MCYT}          & Wacom Intuos A6          & Stylus         & 6.4 inches                     & 2,540 lpi                      & 100 Hz                 \\ \hline
\textbf{BiosecurID}    & Wacom Intuos 3           & Stylus         & 12.5 inches                    & 5,080 lpi                      & 100 Hz                 \\ \hline
\textbf{Biosecure DS2} & Wacom Intuos 3           & Stylus         & 12.5 inches                    & 5,080 lpi                      & 100 Hz                 \\ \hline
\textbf{e-BioSign DS1} & Wacom STU-500            & Stylus         & 5 inches (640x480 pixels)      & 2,540 lpi                      & 200 Hz                 \\
                       & Wacom STU-530            & Stylus         & 5 inches (640x480 pixels)      & 2,540 lpi                      & 200 Hz                 \\
                       & Wacom DTU-1031           & Stylus         & 10.1 inches (1280x800 pixels)  & 2,540 lpi                      & 200 Hz                 \\
                       & Samsung ATIV 7           & Stylus/Finger  & 11.6 inches (1920x1080 pixels) & -                              & -                      \\
                       & Samsung Galaxy Note 10.1 & Stylus/Finger  & 10.1 inches (1280x800 pixels)  & -                              & -                      \\ \hline
\textbf{e-BioSign DS2} & Wacom STU-530            & Stylus         & 5 inches (640x480 pixels)      & 2,540 lpi                      & 200 Hz                 \\
                       & Samsung Galaxy Note 10.1 & Finger         & 10.1 inches (1280x800 pixels)  & -                              & -                      \\
                       & Samsung Galaxy S3        & Finger         & 4.8 inches (720x1280 pixels)   & -                              & -                      \\ \hline
\end{tabular}
\end{table*}

\section{DeepSignDB Database Description}\label{sec_database_description}
The DeepSignDB database comprises a total of 1526 users from four different popular databases (i.e., MCYT, BiosecurID, Biosecure DS2, and e-BioSign DS1) and a novel signature database not presented yet, named e-BioSign DS2. Fig.~\ref{fig:DeepSign_abstract} graphically summarises the design, acquisition devices, and writing tools considered in the DeepSignDB database. A short description of each database regarding the device, writing input, number of acquisition sessions and time gap between them, and type of impostors is included for completeness below. Regarding the type of skilled forgeries, two different approaches can be considered~\cite{2018_HanbookBioAntiSpoofing_signature_Tolosana}: \textit{i) static}, where the forger has access only to the image of the signatures to forge, and \textit{ii) dynamic}, where the forger has access to both the image and also the whole realization process (i.e., dynamics) of the signature to forge. The dynamics can be obtained in the presence of the original writer
or through the use of a video-recording (the case considered in DeepSignDB). We also summarise in Table~\ref{table:devices_features} the main specifications of the acquisition devices considered in each dataset of DeepSignDB. It is important to highlight that for general-purpose devices (the Samsung devices in this case), information about the sensor resolution and sampling rate is not available. In our experience, the sampling rate of general-purpose devices is below 100 Hz and not uniform~\cite{eBioSign_journal}.

\subsection{MCYT}\label{sec_MCYT}
The MCYT database~\cite{Ortega_Garcia2003_MCYT} comprises a total of 25 genuine signatures and 25 skilled forgeries per user, acquired in a single session in blocks of 5 signatures. There are a total of 330 users and signatures were acquired considering a controlled and supervised office-like scenario. Users were asked to sign on a piece of paper, inside a grid that marked the valid signing space, using an inking pen. The paper was placed on a Wacom Intuos A6 USB pen tablet that captured the following time signals: \textit{X} and \textit{Y} spatial coordinates, pressure, pen angular orientations (i.e., azimuth and altitude angles) and timestamps. In addition, pen-up trajectories are available. Regarding the type of impostors, only static forgeries were considered.

\subsection{BiosecurID}\label{sec_BiosecurID}
The BiosecurID database~\cite{Fierrez2009_PAA} comprises a total of 16 genuine signatures and 12 skilled forgeries per user, captured in 4 separate acquisition sessions leaving a two-month interval between them. There are a total of 400 users and signatures were acquired considering a controlled and supervised office-like scenario. Users were asked to sign on a piece of paper, inside a grid that marked the valid signing space, using an inking pen. The paper was placed on a Wacom Intuos 3 pen tablet that captured the following time signals: \textit{X} and \textit{Y} spatial coordinates, pressure, pen angular orientations (i.e., azimuth and altitude angles) and timestamps. Pen-up trajectories are also available. 

Regarding the type of impostors, both static (the first two sessions) and dynamic (the last two sessions) forgeries were considered.

\subsection{Biosecure DS2}\label{sec_BiosecureDS2}
The Biosecure DS2 database~\cite{Ortega09c} comprises a total of 30 genuine signatures and 20 skilled forgeries per user, captured in 2 separate acquisition sessions leaving a three-month time interval between them. There are a total of 650 users and signatures were acquired considering a controlled and supervised office-like scenario. Users were asked to sign on a paper sheet placed on top of a Wacom Intuos 3 device while sitting. The same acquisition conditions were considered as per BiosecurID database.

Regarding the type of impostors, only dynamic forgeries were considered.

\subsection{e-BioSign DS1}\label{sec_eBioSign_DS1}
The e-BioSign DS1 database~\cite{eBioSign_journal} is composed of five different devices. Three of them are specifically designed for capturing handwritten data (i.e., Wacom STU-500, STU-530, and DTU-1031), while the other two are general-purpose tablets not designed for that specific task (Samsung ATIV 7 and Galaxy Note 10.1). It is worth noting that all five devices were used with their own pen stylus. Additionally, the two Samsung devices were used with the finger as input, allowing the analysis of the writing input on the system performance. The same capturing protocol was used for all five devices: devices were placed on a desktop and subjects were able to rotate them in order to feel comfortable with the writing position. The software for capturing handwriting and signatures was developed in the same way for all devices in order to minimise the variability of the user during the acquisition process. 

Signatures were collected in two sessions for 65 subjects with a time gap between sessions of at least 3 weeks. For each user and writing input, there are a total of 8 genuine signatures and 6 skilled forgeries. For the case of using the stylus as input, information related to \textit{X} and \textit{Y} spatial coordinates, pressure and timestamp is recorded for all devices. In addition, pen-up trajectories are also available. However, pressure information and pen-up trajectories are not recorded when the finger is used as input.

Regarding the type of impostors, both dynamic and static forgeries were considered in the first and second acquisition sessions, respectively.

\begin{table*}[t]
\caption{Experimental protocol details of the DeepSignDB evaluation dataset (442 users). Numbers are per user and device.}
\label{table:experimental_protocol}
\centering
\scalebox{0.87}{
\begin{tabular}{ccccccc}
\multicolumn{7}{c}{\textbf{STYLUS WRITING INPUT}}                                                                                                                                                                                                                                                                           \\ \hline
\textbf{Database}    & \textbf{\#Users}     & \textbf{Devices}                                                                                                            & \textbf{\#Train Genuine Signatures} & \textbf{\#Test Genuine Signatures} & \textbf{\#Test Skilled Forgeries} & \textbf{\#Test Random Forgeries} \\ \hline
MCYT                 & 100                  & Wacom Intuos A6                                                                                                             & 1/4 (Session 1)                     & 21 (rest)                          & 25 (all)                          & 99 (one of the rest users)       \\ \hline
BiosecurID           & 132                  & Wacom Intuos 3                                                                                                              & 1/4 (Session 1)                     & 12 (Sessions 2-4)                  & 12 (all)                          & 131 (one of the rest users)      \\ \hline
Biosecure DS2        & 140                  & Wacom Intuos 3                                                                                                              & 1/4 (Session 1)                     & 15 (Session 2)                     & 20 (all)                          & 139 (one of the rest users)      \\ \hline
e-BioSign DS1           & 35                   & \begin{tabular}[c]{@{}c@{}}W1: Wacom STU-500\\ W2: Wacom STU-530\\ W3: Wacom DTU-1031\\ W4: Samsung ATIV 7\\ W5: Samsung Note 10.1\end{tabular} & 1/4 (Session 1)                     & 4 (Session 2)                      & 6 (all)                           & 34 (one of the rest users)       \\ \hline
e-BioSign DS2          & 35                   & W2: Wacom STU-530                                                                                                               & 1/4 (Session 1)                     & 4 (Session 2)                      & 6 (all)                           & 34 (one of the rest users)       \\ \hline
\multicolumn{1}{l}{} & \multicolumn{1}{l}{} & \multicolumn{1}{l}{}                                                                                                        & \multicolumn{1}{l}{}                & \multicolumn{1}{l}{}               & \multicolumn{1}{l}{}              & \multicolumn{1}{l}{}             \\
\multicolumn{7}{c}{\textbf{FINGER WRITING INPUT}}                                                                                                                                                                                                                                                                           \\ \hline
\textbf{Database}    & \textbf{\#Users}     & \textbf{Devices}                                                                                                            & \textbf{\#Train Genuine Signatures} & \textbf{\#Test Genuine Signatures} & \textbf{\#Test Skilled Forgeries} & \textbf{\#Test Random Forgeries} \\ \hline
e-BioSign DS1           & 35                   & \begin{tabular}[c]{@{}c@{}}W4: Samsung ATIV 7\\ W5: Samsung Note 10.1\end{tabular}                                                  & 1/4 (Session 1)                     & 4 (Session 2)                      & 6 (all)                           & 34 (one of the rest users)       \\ \hline
e-BioSign DS2          & 35                   & \begin{tabular}[c]{@{}c@{}}W5: Samsung Note 10.1\\ W6: Samsung S3\end{tabular}                                                      & 1/4 (Session 1)                     & 4 (Session 2)                      & 6 (all)                           & 34 (one of the rest users)       \\ \hline
\end{tabular}
}
\end{table*}

\subsection{e-BioSign DS2}\label{sec_newDataSet}
DeepSignDB database also includes a new on-line signature dataset not presented yet, named e-BioSign DS2. This dataset follows the same capturing protocol as e-BioSign DS1. Three different devices were considered: a Wacom STU-530 specifically designed for capturing handwritten data, a Samsung Galaxy Note 10.1 general-purpose tablet, and a Samsung Galaxy S3 smartphone. For the first device, signatures where captured using the stylus in an office-like scenario, i.e., the device was placed on a desktop and subjects were able to rotate it in order to feel comfortable with the writing position. For the Samsung Galaxy Note 10.1 tablet and Galaxy S3 smartphone, the finger was used as input. The acquisition conditions emulated a mobile scenario where users had to sign while sitting. 

Signatures were collected in two sessions for 81 users with a time gap between sessions of at least 3 weeks. For each user, device, and writing input, there are a total of 8 genuine signatures and 6 skilled forgeries. For the case of using the stylus as input, information related to \textit{X} and \textit{Y} spatial coordinates, pressure and timestamp is recorded for all devices. In addition, pen-up trajectories are also available. However, pressure information and pen-ups trajectories are not recorded when the finger is used as input.

Regarding the type of impostors, only dynamic forgeries were considered. 

\section{DeepSignDB Benchmark}\label{sec_deepSign_benchmark}
This section reports the benchmark evaluation carried out for the DeepSignDB on-line handwritten signature database. Sec.~\ref{sec_experimentalProtocol} describes all the details of our proposed standard experimental protocol to be used for the research community in order to facilitate the fair comparison of novel approaches with the state of the art. Finally, Sec.~\ref{sec_experimentalResults} analyses the results achieved using our proposed TA-RNN system and compares it with the preliminary benchmark results achieved in~\cite{Tolosana_2019_DeepSignDB_ICDAR}, based on a robust DTW and RNN two-stage approach~\cite{2018_IEEEAccess_RNN_Tolosana}.

\subsection{Experimental Protocol}\label{sec_experimentalProtocol}
The DeepSignDB database has been divided into two different datasets, one for the development and training of the systems and the other one for the final evaluation. The development dataset comprises around 70\% of the users of each database whereas the remaining 30\% are included in the evaluation dataset. It is important to note that each dataset comprises different users in order to avoid biased results. 

For the training of the systems, the development dataset comprises a total of 1084 users. In our experiments, we have divided this dataset into two different subsets, training (80\%) and validation (20\%). However, as this dataset is used only for development, and not for the final evaluation of the systems, we prefer not to set any restriction and let researchers use it as they like.

For the final testing of the systems, the remaining 442 users of the DeepSignDB database are included in the evaluation dataset in order to perform a complete analysis of the signature verification systems, and see their generalisation capacity to different scenarios. The following aspects have been considered in the final experimental protocol design:
\begin{itemize}
\item \textbf{Inter-session variability:} genuine signatures from different sessions are considered for training and testing (different acquisition blocks for the MCYT database).
\item \textbf{Number of training signatures:} two different cases are considered, the case of having just one genuine signature from the first session (1vs1) and the case of using the first 4 genuine signatures from the first session (4vs1). In this study the final score of the 4vs1 case is obtained as the average score of the 4 one-to-one comparisons.
\item \textbf{Impostor scenario:} skilled and random forgeries are considered in the experimental protocol. For the skilled forgery case, all available skilled forgery samples are included in the analysis whereas for the random forgery case, one genuine sample of each of the remaining users of the same database is considered. This way verification systems are tested with different types of presentation attacks~\cite{2018_HanbookBioAntiSpoofing_signature_Tolosana}. 
\item \textbf{Writing input:} stylus and finger scenarios are also considered in the experimental protocol due to the high acceptance of the society to use mobile devices on a daily basis~\cite{MobileAddictive}.
\item \textbf{Acquisition device:} eight different devices are considered in the experimental protocol. This will allow to measure the generalisation capacity of the proposed system to different acquisition devices that can be found in different applications.
\end{itemize}

Table~\ref{table:experimental_protocol} describes all the experimental protocol details of the DeepSignDB evaluation dataset for both stylus (top) and finger (bottom) writing inputs.

\subsection{Experimental Results}\label{sec_experimentalResults}
Two different scenarios are evaluated in our proposed standard experimental protocol. First, an office-like scenario where users perform their signatures using the stylus as input (Table~\ref{table:experimental_protocol}, top), and then a mobile scenario where users perform their signatures using the finger on mobile general-purpose devices (Table~\ref{table:experimental_protocol}, bottom). It is important to remark that the DeepSignDB results are obtained after performing all the signature comparisons of the corresponding databases together, and not through the average EERs of the corresponding databases. This way we consider a single system threshold, simulating real scenarios.

\begin{table*}[t]
\caption{System performance results (EER) over the DeepSignDB evaluation dataset. \textbf{Stylus} scenario.}
\label{table:stylus_results}
\centering
\scalebox{1.}{
\begin{tabular}{c|ccc|ccc|l|cc|cc|}
\multicolumn{1}{l|}{} & \multicolumn{6}{c|}{\textbf{Skilled Forgeries}}                                                          & \multicolumn{1}{c|}{} & \multicolumn{4}{c|}{\textbf{Random Forgeries}}                                                           \\ \cline{2-7} \cline{9-12} 
                      & \multicolumn{3}{c|}{\textbf{1 Training Signature}} & \multicolumn{3}{c|}{\textbf{4 Training Signatures}} &                       & \multicolumn{2}{c|}{\textbf{1 Training Signature}} & \multicolumn{2}{c|}{\textbf{4 Training Signatures}} \\ \cline{2-7} \cline{9-12} 
                      & DTW          & RNNs         & TA-RNNs              & DTW           & RNNs         & TA-RNNs              &                       & DTW                      & TA- RNNs                & DTW                      & TA-RNNs                  \\ \cline{1-7} \cline{9-12} 
MCYT                  & 9.1          & 10.5         & \textbf{4.4}         & 7.2           & 10.1         & \textbf{4.3}         &                       & 1.2                      & \textbf{1.1}            & 0.6                      & \textbf{0.2}             \\ \cline{1-7} \cline{9-12} 
BiosecurID            & 8.1          & 3.9          & \textbf{1.9}         & 6.5           & 3.4          & \textbf{1.3}         &                       & 1.0                      & \textbf{0.6}            & 0.6                      & \textbf{0.1}             \\ \cline{1-7} \cline{9-12} 
Biosecure DS2         & 14.2         & 8.0          & \textbf{4.2}         & 12.1          & 7.4          & \textbf{3.0}         &                       & 2.5                      & \textbf{1.9}            & 1.6                      & \textbf{1.1}             \\ \cline{1-7} \cline{9-12} 
eBS DS1 w1            & 15.3         & 11.4         & \textbf{5.4}         & 9.3           & 9.0          & \textbf{4.3}         &                       & 3.2                      & \textbf{2.5}            & 0.7                      & \textbf{0.1}             \\
eBS DS1 w2            & 12.0         & 8.2          & \textbf{4.0}         & 11.4          & 7.1          & \textbf{2.9}         &                       & \textbf{1.3}             & 1.7                     & \textbf{0.7}             & 1.4                      \\
eBS DS1 w3            & 14.5         & 14.3         & \textbf{5.4}         & 12.1          & 11.4         & \textbf{4.8}         &                       & \textbf{0.9}             & 1.6                     & \textbf{0.3}             & 0.4                      \\
eBS DS1 w4            & 14.6         & 13.2         & \textbf{5.8}         & 11.4          & 12.1         & \textbf{5.2}         &                       & \textbf{1.1}             & 1.4                     & \textbf{0.7}             & 0.9                      \\
eBS DS1 w5            & 14.9         & 18.9         & \textbf{10.6}        & 12.9          & 14.0         & \textbf{8.0}         &                       & \textbf{2.7}             & 4.1                     & 2.1                      & \textbf{1.4}             \\ \cline{1-7} \cline{9-12} 
eBS DS2 w2            & 9.6          & 3.9          & \textbf{3.7}         & 8.3           & 2.9          & \textbf{2.8}         &                       & 2.7                      & \textbf{2.2}            & \textbf{0.7}             & 0.9                      \\ \cline{1-7} \cline{9-12} \hline \hline
DeepSignDB            & 11.2          & 8.5          & \textbf{4.2}         & 9.3           & 7.9          & \textbf{3.3}         &                       & 1.8                      & \textbf{1.5}            & 1.1             & \textbf{0.6}                      \\ \cline{1-7} \cline{9-12}
\end{tabular}
}
\end{table*}

\begin{table*}[t]
\caption{System performance results (EER) over the DeepSignDB evaluation dataset. \textbf{Finger} scenario.}
\label{table:finger_results}
\centering
\scalebox{1.}{
\begin{tabular}{c|ccc|ccc|l|cc|cc|}
\multicolumn{1}{l|}{} & \multicolumn{6}{c|}{\textbf{Skilled Forgeries}}                                                          & \multicolumn{1}{c|}{} & \multicolumn{4}{c|}{\textbf{Random Forgeries}}                                                           \\ \cline{2-7} \cline{9-12} 
                      & \multicolumn{3}{c|}{\textbf{1 Training Signature}} & \multicolumn{3}{c|}{\textbf{4 Training Signatures}} &                       & \multicolumn{2}{c|}{\textbf{1 Training Signature}} & \multicolumn{2}{c|}{\textbf{4 Training Signatures}} \\ \cline{2-7} \cline{9-12} 
                      & DTW          & RNNs         & TA-RNNs              & DTW          & RNNs         & TA-RNNs               &                       & DTW                      & TA- RNNs                & DTW                      & TA-RNNs                  \\ \cline{1-7} \cline{9-12} 
eBS DS1 w4            & 20.0         & 20.7         & \textbf{18.8}        & 19.3         & 19.3         & \textbf{16.6}         &                       & \textbf{0.7}             & 1.0                     & \textbf{0.7}             & \textbf{0.7}             \\
eBS DS1 w5            & 20.2         & 21.0         & \textbf{16.4}        & 16.4         & 20.0         & \textbf{13.3}         &                       & \textbf{1.7}             & \textbf{1.7}            & 1.4                      & \textbf{0.7}             \\ \cline{1-7} \cline{9-12} 
eBS DS2 w5            & 14.5         & 17.0         & \textbf{9.8}         & 12.6         & 16.9         & \textbf{10.0}         &                       & \textbf{0.6}             & 2.3                     & \textbf{0.2}             & 1.4                      \\
eBS DS2 w6            & 12.8         & 13.6         & \textbf{8.4}         & 12.1         & 13.6         & \textbf{5.7}          &                       & \textbf{1.3}             & 1.7                     & \textbf{0.8}             & 1.4                      \\ \cline{1-7} \cline{9-12} \hline \hline
DeepSignDB            & 16.6         & 18.6         & \textbf{13.8}         & 14.8         & 17.3         & \textbf{11.3}          &                       & \textbf{1.2}             & 1.8                     & \textbf{0.7}             & 1.0                      \\ \cline{1-7} \cline{9-12}

\end{tabular}
}
\end{table*}

\subsubsection{\textbf{Stylus Writing Input Scenario}}\label{sec_stylusWritingInput}
For the development of the systems, only signatures acquired using the stylus are considered, ending up with around 309K genuine and impostor comparisons (247K and 62K for training and validation, respectively). It is important to remark that: \textit{i)} the same number of genuine and impostor comparisons are used in order to avoid bias, and \textit{ii)} both skilled and random forgeries are used as impostors during the development process in order to provide robust systems against both types of attacks.

Table~\ref{table:stylus_results} depicts the evaluation performance results of our proposed TA-RNN approach for the whole DeepSignDB evaluation dataset and for each of the datasets included in it when using the stylus as input. In addition, we compare the proposed TA-RNNs with the preliminary benchmark results presented in~\cite{Tolosana_2019_DeepSignDB_ICDAR} for completeness. In that study, RNNs outperformed DTW for skilled forgeries. However, for random forgeries, DTW further outperformed RNNs with very low EERs. Therefore, random forgery results for RNNs are not shown in Table~\ref{table:stylus_results} in order to avoid meaningless results.

Both RNN and TA-RNN systems have been implemented under Keras framework using Tensorflow as back-end, with a NVIDIA GeForce RTX 2080 Ti GPU. The weights of the BGRU and feed-forward layers are initialised by random
values drawn from the zero-mean Gaussian distribution with
standard deviation 0.05. Adam optimiser is considered with
default parameters (learning rate of 0.001) and a loss function based on binary cross-entropy. It is worth mentioning that in average a single one-to-one signature comparison took 0.72 seconds, making it feasible for real time applications. On the other hand, the training of the deep learning models took around 48 hours. Note these times can be significantly reduced with higher performance computing.

Analysing skilled forgeries, our proposed TA-RNN approach outperforms in large margins previous approaches. For the scenario of considering just 1 training signature per user, TA-RNNs achieves an absolute improvement of 7.0\% and 4.3\% EERs compared with the DTW and RNN systems, respectively. It is important to remark that we are just training one model for the whole DeepSignDB development dataset, and not one specific model per dataset. Our proposed writer-independent TA-RNN approach shows a high ability to generalise well along different scenarios, users, and devices, achieving EERs even below 2.0\% in challenging scenarios where dynamic skilled forgery impostors and just one training signature per user are considered. Similar results are obtained for the scenario of increasing the number of training signatures to 4. TA-RNNs achieves an absolute improvement of 6.0\% and 4.6\% EERs compared with the DTW and RNN systems, respectively. 

We now analyse the random forgery results of Table~\ref{table:stylus_results}. In general, similar results are observed among the DTW and TA-RNNs. For the case of using just 1 training signature, our proposed TA-RNNs is able to outperform the robust DTW in 5 out of 9 different datasets, achieving a final 1.5\% EER for the whole DeepSignDB evaluation dataset, an absolute improvement of 0.3\% EER compared with the DTW system. This result improves further when we increase the number of training signatures to 4, with EERs very low. 

\begin{figure*}[t]
\centering
\begin{subfigure}[tb]{0.39\textwidth}
\centering
\centerline{\includegraphics[width=\linewidth]{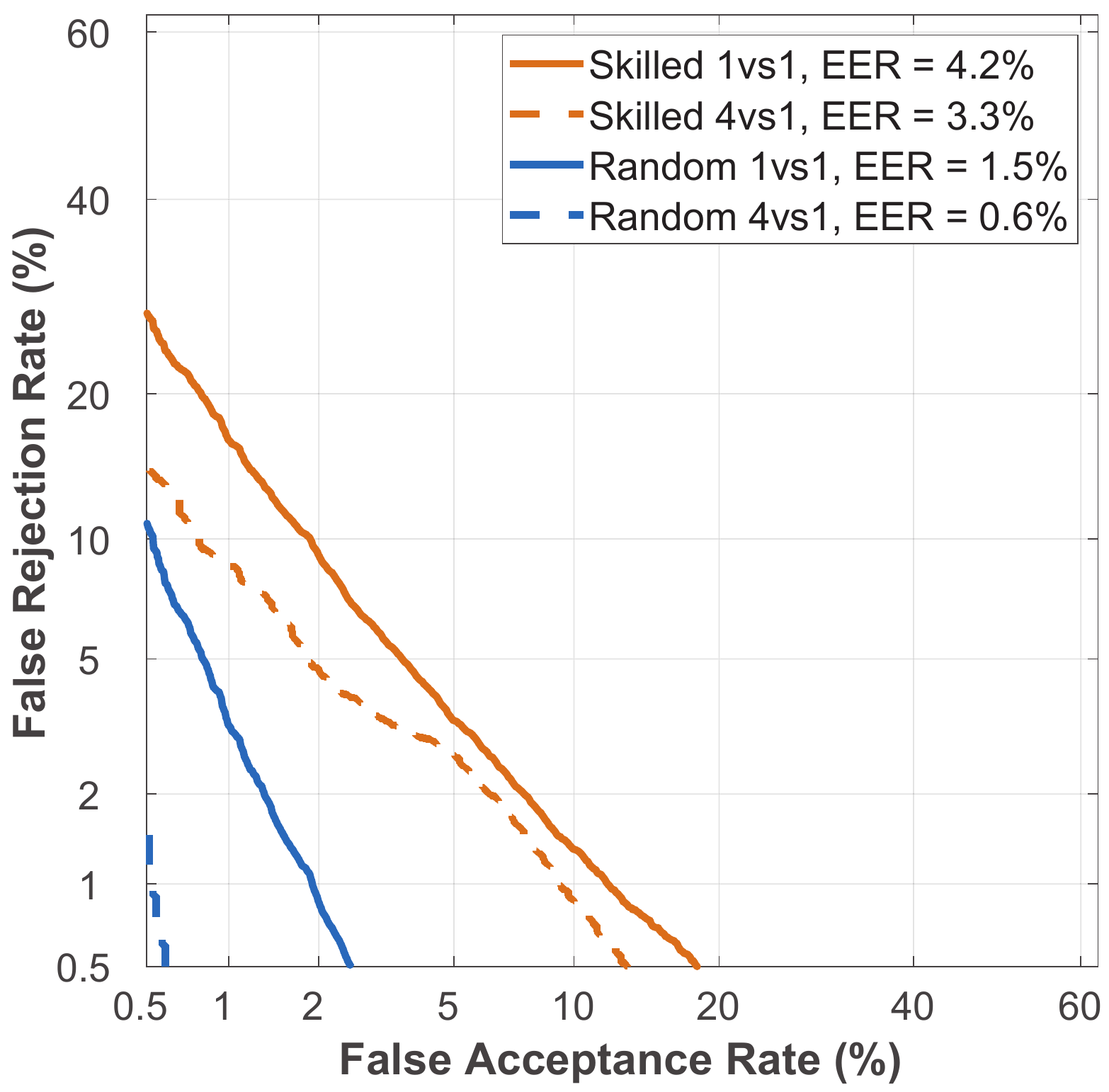}}
\caption{Stylus} \label{fig:DET_stylus}
\end{subfigure}
\begin{subfigure}[tb]{0.39\textwidth}
\centerline{\includegraphics[width=\linewidth]{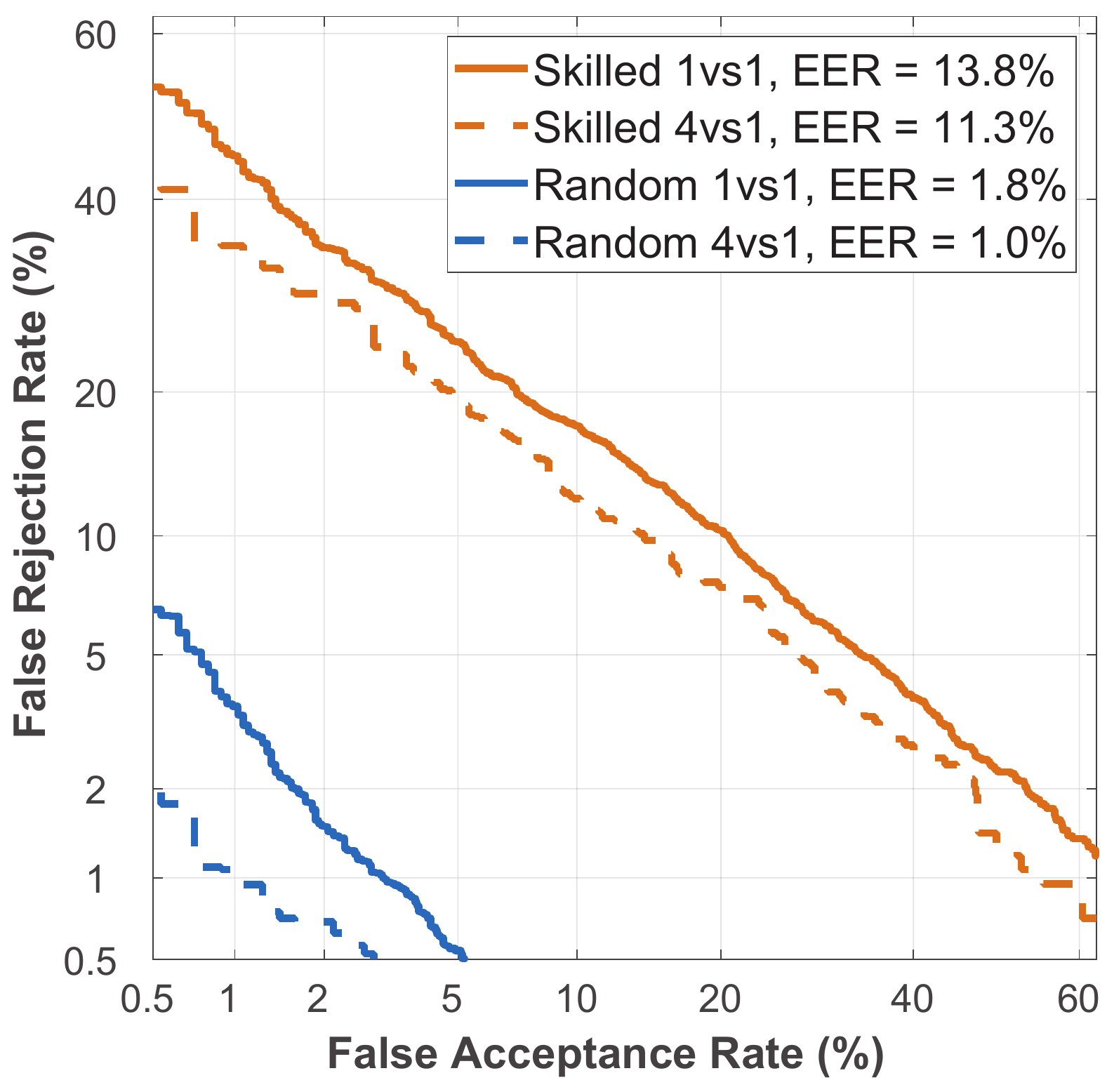}}
\caption{Finger} \label{fig:DET_finger}
\end{subfigure}
\caption{System performance results of our proposed TA-RNN system over the DeepSignDB evaluation dataset.} \label{fig:DET_stylus_fingerº}\vspace{-0.4cm}
\end{figure*}

Finally, Fig.~\ref{fig:DET_stylus} depicts the DET curve of the TA-RNN performance results obtained using the whole DeepSignDB evaluation dataset for the stylus scenario, for completeness. The results achieved put in evidence the success of our proposed TA-RNN approach, obtaining very good results against both skilled and random forgeries, and overcoming the original training problems described in~\cite{2018_IEEEAccess_RNN_Tolosana}.

\subsubsection{\textbf{Finger Writing Input Scenario}}\label{sec_FingerWritingInput}
We consider the same on-line signature verification systems trained in the previous section for the case of using the stylus as input. This way we can: \textit{i)} evaluate the generalisation capacity of the network to unseen writing inputs, i.e., the finger, and \textit{ii)} encourage all the research community to use DeepSignDB and explore new methods such as transfer learning in this challenging scenario where the number of public databases is very scarce~\cite{pan2010survey,hu2015deep}. As pressure information (and its derivative) is not available on the finger scenario, these time functions are set to zero when using the stylus system.

Table~\ref{table:finger_results} depicts the evaluation performance results of our proposed TA-RNN approach for both the whole DeepSignDB dataset and for each of the databases included in it when using the finger as input. Analysing skilled forgeries, our proposed TA-RNNs outperforms DTW and RNNs. For the scenario of considering just 1 training signature per user, TA-RNNs achieves an absolute improvement of 2.8\% and 4.8\% EERs compared with the DTW and RNN systems, respectively. Similar trends are observed when increasing the number of training signatures to 4. Analysing random forgeries, the DTW system slightly outperforms the proposed TA-RNN system, achieving both very low EERs for the case of using 1 or 4 training signatures per user.

Finally, Fig.~\ref{fig:DET_finger} depicts the DET curve of the TA-RNN performance results obtained using the whole DeepSignDB evaluation dataset for the finger scenario. Analysing skilled forgeries, we can observe a high degradation of the system performance compared with the stylus scenario. Concretely, absolute worsening of 9.6\% and 8.0\% EERs for the scenarios of using 1 and 4 training signatures, respectively. These results agree with preliminary studies in the field~\cite{eBioSign_journal,Lai_TIFS_2018}. Therefore, we encourage the research community to put their efforts in this challenging but important scenario.

\section{Conclusions}\label{conclusions}
This article has presented the DeepSignDB on-line handwritten signature database, the largest on-line signature database to date. This database comprises more than 70K signatures acquired using both stylus and finger inputs from a total of 1526 users. Two acquisition scenarios are considered (i.e., office and mobile), with a total of 8 different devices. Additionally, different types of impostors and number of acquisition sessions are considered along the database.

In addition, we have proposed a standard experimental protocol and benchmark to be used for the research community in order to perform a fair comparison of novel approaches with the state of the art. Finally, we have adapted and evaluated our recent deep learning approach named Time-Aligned Recurrent Neural Networks (TA-RNNs) for on-line handwritten signature verification, which combines the potential of Dynamic Time Warping and Recurrent Neural Networks to train more robust systems against forgeries.

Our proposed TA-RNN system has further outperformed all previous state-of-the-art approaches, achieving results even below 2.0\% EER for some datasets of DeepSignDB when considering skilled forgery impostors and just one training signature per user. The results achieved put in evidence the high ability of our proposed approach to generalise well along different scenarios, users, and acquisition devices.

For future work, we encourage the research community to use DeepSignDB database for several purposes: \textit{i)} perform a fair comparison of novel approaches with the state of the art (we refer the reader to download the DeepSignDB\footnote{\url{https://github.com/BiDAlab/DeepSignDB}} and follow the ICDAR 2021 Competition on On-Line Signature Verification, SVC 2021\footnote{\url{https://sites.google.com/view/SVC2021/home}}) \textit{ii)} evaluate the limits of novel DL architectures, and \textit{iii)} carry out a more exhaustive analysis of the challenging finger input scenario. In addition, DeepSignDB can be also very useful to study neuromotor aspects related to handwriting and touchscreen interaction~\cite{2019_BookLogNormal_ComplexSignTouch_Vera} across population groups and age~\cite{2018_IETB_DetectChildTouch_Acien} for diverse applications like e-learning and e-health~\cite{2020_Signature_CognitiveComputation}. Finally, we plan to evaluate the usability and performance improvement of our proposed TA-RNN approach for other signature verification approaches based on the use of synthetic samples~\cite{2021_AAAI_DeepWriteSYN,galbally09ICDARenrollment}, and for other behavioral biometric traits such as keystroke biometrics~\cite{2020_SDIM_KeysPandemic_Morales}.

\section*{Acknowledgments}
This work has been supported by projects: PRIMA (H2020-MSCA-ITN-2019-860315), TRESPASS-ETN (H2020-MSCA-ITN-2019-860813), BIBECA (RTI2018-101248-B-I00 MINECO/FEDER), Bio-Guard (Ayudas Fundaci\'on BBVA a Equipos de Investigaci\'on Cient\'ifica 2017), and by UAM-Cecabank. R. Tolosana is supported
by Comunidad de Madrid and Fondo Social Europeo. Spanish Patent Application (P202030060).

{
\bibliographystyle{IEEEtran}
\bibliography{egbib2}
}

\begin{IEEEbiography}[{\includegraphics[width=1in,height=1.25in,clip,keepaspectratio]{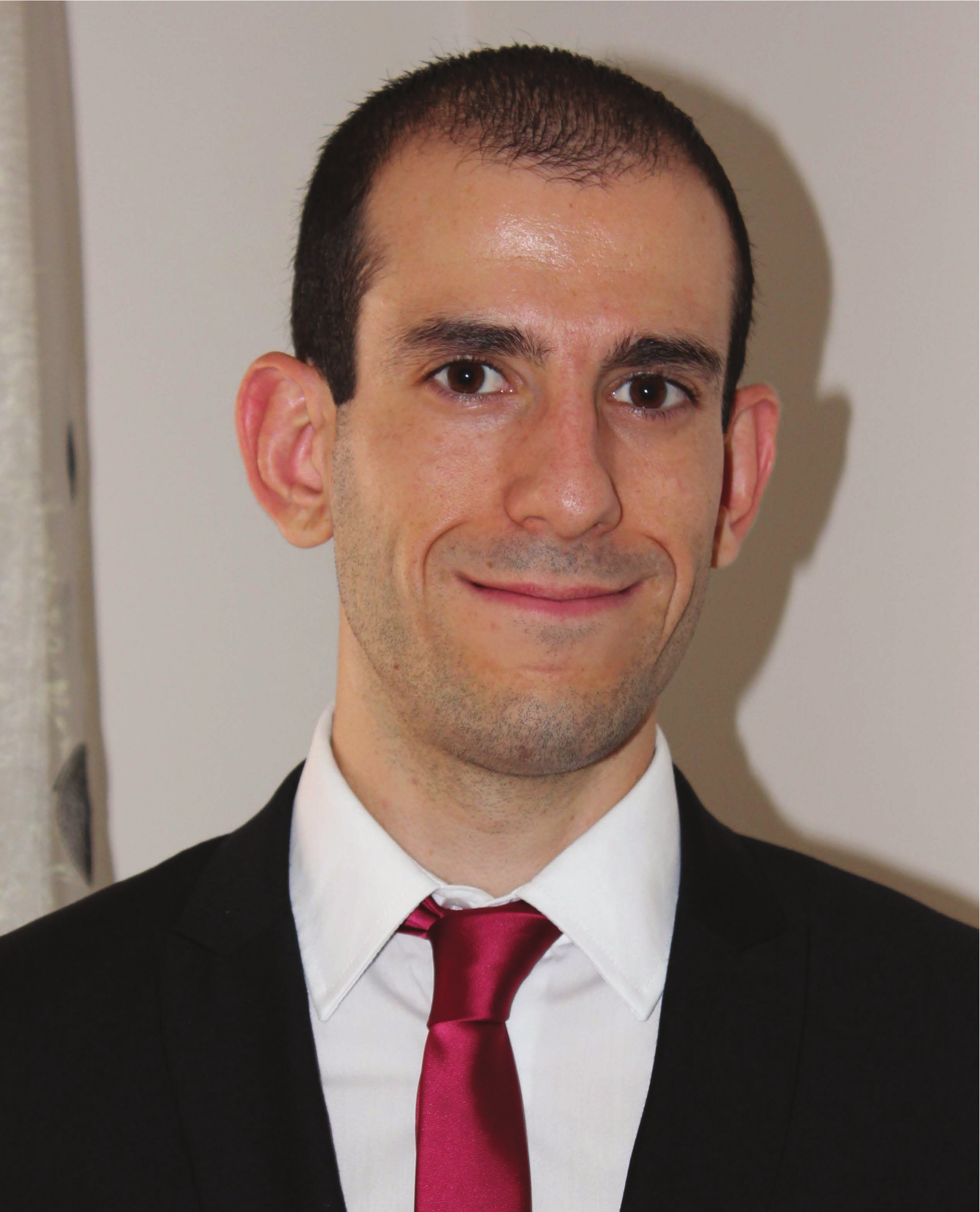}}]%
{Ruben Tolosana}
received the M.Sc. degree in Telecommunication Engineering, and his Ph.D. degree in Computer and Telecommunication Engineering, from Universidad Autonoma de Madrid, in 2014 and 2019, respectively. In 2014, he joined the Biometrics and Data Pattern Analytics - BiDA Lab at the Universidad Autonoma de Madrid, where he is currently collaborating as a PostDoctoral researcher. Since then, Ruben has been granted with several awards such as the FPU research fellowship from Spanish MECD (2015), and the European Biometrics Industry Award (2018). His research interests are mainly focused on signal and image processing, pattern recognition, and machine learning, particularly in the areas of DeepFakes, HCI, and Biometrics. He is author of several publications and also collaborates as a reviewer in high-impact conferences (WACV, ICPR, ICDAR, IJCB, etc.) and journals (IEEE TPAMI, TCYB, TIFS, TIP, ACM CSUR, etc.). Finally, he is also actively involved in several National and European projects.
\end{IEEEbiography}

\begin{IEEEbiography}[{\includegraphics[width=1in,height=1.25in,clip,keepaspectratio]{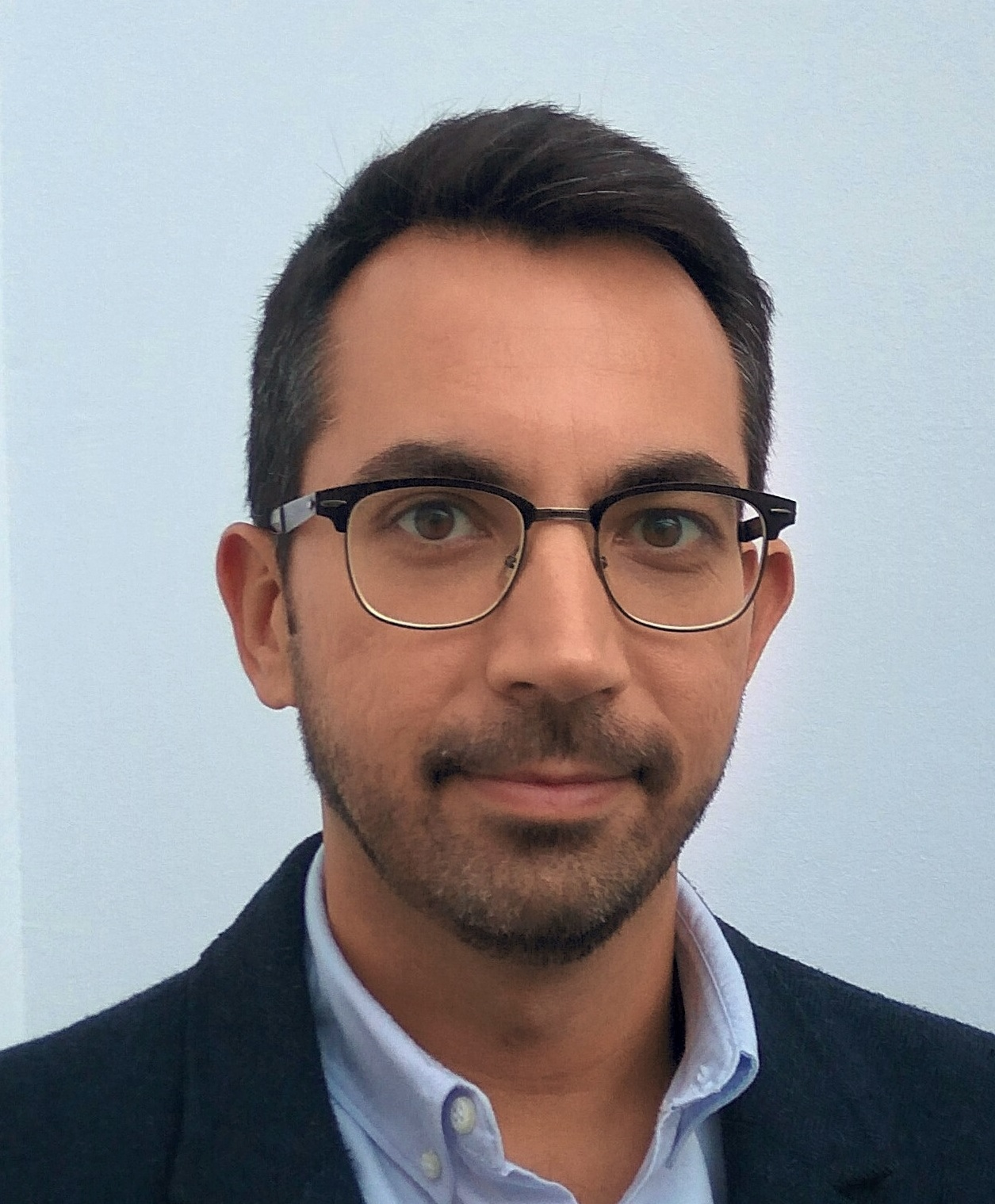}}]{Ruben Vera-Rodriguez} received the M.Sc. degree in telecommunications engineering from Universidad de Sevilla, Spain, in 2006, and the Ph.D. degree in electrical and electronic engineering from Swansea University, U.K., in 2010. Since 2010, he has been affiliated with the Biometric Recognition Group, Universidad Autonoma de Madrid, Spain, where he is currently an Associate Professor since 2018. His research interests include signal and image processing, pattern recognition, and biometrics, with emphasis on signature, face, gait verification and forensic applications of biometrics. He is actively involved in several National and European projects focused on biometrics. Ruben has been Program Chair for the IEEE 51st International Carnahan Conference on Security and Technology (ICCST) in 2017; and the 23rd Iberoamerican Congress on Pattern Recognition (CIARP 2018) in 2018.
\end{IEEEbiography}

\begin{IEEEbiography}[{\includegraphics[width=1in,height=1.25in,clip,keepaspectratio]{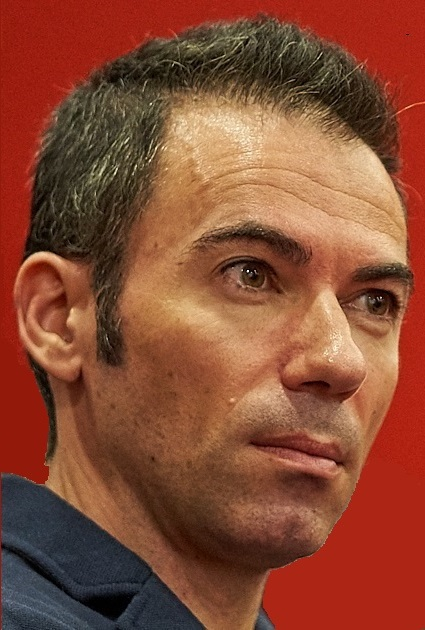}}]{Julian Fierrez} received the M.Sc. and Ph.D. degrees in telecommunications engineering from the Universidad Politecnica de Madrid, Spain, in 2001 and 2006, respectively. Since 2004 he has been at Universidad Autonoma de Madrid, where he is currently an Associate Professor. From 2007 to 2009 he was a Visiting Researcher with Michigan State University, USA, under a Marie Curie postdoc. His research is on signal and image processing, HCI, responsible AI, and biometrics for security and human behavior analysis. He is actively involved in large EU projects in these topics (e.g., TABULA RASA and BEAT in the past, now IDEA-FAST and TRESPASS-ETN), and has attracted notable impact for his research. He was a recipient of a number of distinctions, including the EAB Industry Award 2006, the EURASIP Best Ph.D. Award 2012, and the 2017 IAPR Young Biometrics Investigator Award. He has received best paper awards at ICB and ICPR. He is Associate Editor of the IEEE TRANSACTIONS ON INFORMATION FORENSICS AND SECURITY and the IEEE TRANSACTIONS ON IMAGE PROCESSING. He is member of the ELLIS Society.
\end{IEEEbiography}

\begin{IEEEbiography}[{\includegraphics[width=1in,height=1.25in,clip,keepaspectratio]{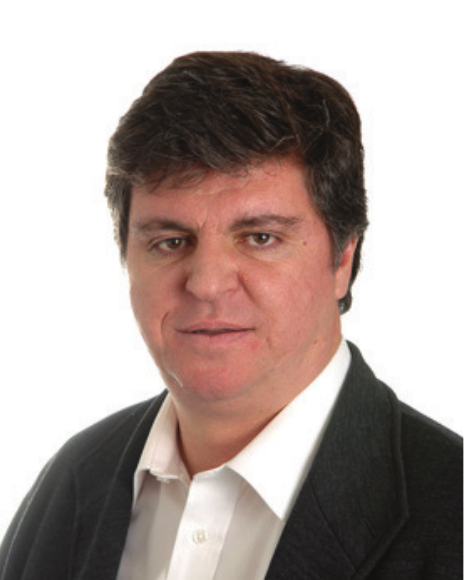}}]%
{Javier Ortega-Garcia}
received the M.Sc. degree in electrical engineering and the Ph.D. degree (cum laude) in electrical engineering from Universidad Politecnica de Madrid, Spain, in 1989 and 1996, respectively. He is currently a Full Professor at the Signal Processing Chair in Universidad Autonoma de Madrid - Spain, where he holds courses on biometric recognition and digital signal processing. He is a founder and Director of the BiDA-Lab, Biometrics and Data Pattern Analytics Group. He has authored over 300 international contributions, including book chapters, refereed journal, and conference papers. His research interests are focused on biometric pattern recognition (on-line signature verification, speaker recognition, human-device interaction) for security, e-health and user profiling applications. He chaired Odyssey-04, The Speaker Recognition Workshop, ICB-2013, the 6th IAPR International Conference on Biometrics, and ICCST2017, the 51st IEEE International Carnahan Conference on Security Technology.
\end{IEEEbiography}

\end{document}